# HIGH PERFORMANCE CONTROLLERS FOR SPEED AND POSITION INDUCTION MOTOR DRIVE USING NEW REACHING LAW

Salah Eddine Rezgui[1], Hocine Benalla[1]

[1]Constantine Electrotechnics Laboratory, Engineer Sciences Faculty of Mentouri's Constantine University, ALGERIA
r_salaheddine1@yahoo.fr
benalladz@yahoo.fr

*ABSTRACT*

*This paper present new approach in robust indirect rotor field oriented (IRFOC) induction motor (IM) control. The introduction of new exponential reaching law (ERL) based sliding mode control (SMC) improve significantly the performances compared to the conventional SMC which are well known susceptible to the annoying chattering phenomenon, so, the elimination of the chattering is achieved while simplicity and high performance speed and position tracking are maintained. Simulation results are given to discuss the performances of the proposed control method.*

## 1. INTRODUCTION

The field-oriented control technique has been widely used when high-performance rotary machine drive is required, especially the indirect field oriented control (IFOC) which is one of the most effective vector control of IM due to the simplicity of designing and implementation [1]. Advent of high switching frequency PWM inverters has made it possible to apply sophisticated control strategies to AC motor drives. The space vector modulation (SVM) technique has become one of the most important PWM methods. It appears to be the best alternative for a three phase switching power converter because it provides an optimization of converter operation; reducing the commutations of the power semiconductor, does not generate subharmonic components, with capacity about 90.6% of DC link exploitation, and it becomes the best technique to reduce the ripple in the torque signal [2].

However the classical proportional-integral and derivative (PID) controller's which are the main control tool being used in AC machine drives, have major drawbacks that are the sensitivity to the system-parameters variations and bad rejection of external disturbances. To surmount these drawbacks and improve the induction motor control techniques, a set of papers are presented in the literature, we cite as non exhaustive examples; the fuzzy and neuro-fuzzy control [3,4], the neural network control [5], the sliding mode control [6], and the neuro-fuzzy-sliding mode control [7].

All of aforementioned techniques present a good performances, but in term of fast response and robustness against uncertainties including; parametric variations, external disturbance rejection, and unmodelled dynamics, the sliding mode controllers is computationally simple compared to adaptive controllers with parameter estimation and have the advantage and best high results that leads to the improvement of IM control [8].

The SMC was firstly investigated for electric motors by Utkin [9], its design method is generally based on two steps, first, the selection of an appropriate sliding surface, and second, the synthesis of a control law such that a reaching condition is satisfied which makes the selected surface attractive, then, that evolves two modes; reaching mode and sliding mode. Nevertheless, before attaining the sliding surface (i.e. reaching mode), the system cannot dominate the variations in parameters and external disturbances, thus making weak robustness of the system [10].

It is a well known opinion that the major drawback of sliding mode control is the chattering phenomenon. The chattering consists of the oscillation of the control signal, tied to the discontinuous nature of the control law, at a frequency and amplitude which cannot be tolerated in some practical applications, especially mechanical and electro-mechanical ones, also this harmful phenomenon, is caused by unmodeled dynamics or discrete time implementation [11]. A large number of works deals to reduce the chattering have been developed, first of them consist of a soft continuous approximation of the discontinuous law, when boundary layer of definite width on both sides of sliding plane is introduced [12], This method can give a chattering reduction system but a finite steady state error must exist, other works introduce new switching surfaces; as nonlinear and time-varying parabolic sliding surface to improve the discontinuous control law part of a classical sliding mode controller [13], also in [14], the proposed modification consists in modifying the sliding surface using fuzzy rules. These structures make the implementation of controller more complex and affect system convergence to the surface.

The approach, called higher order SMC, has been proposed in order to reduce the chattering phenomenon. Instead of influencing the first sliding variable time derivative, the *signum* function acts on its higher order time derivative, then the chattering problem is avoided by removing discontinuity from control input. The papers proposed second, and high order controllers respectively in [15,16] but, in spite of chattering attenuation, such methods are overly complex and difficult to implement in practice.

Besides of these works, in last decade a new concept called Reaching Law Method (RLC) had been introduced in sliding mode control [17]. In this method the error dynamics is specified in the reaching mode, with selecting appropriately parameters both the dynamic quality of the SMC system can be controlled and the chattering is removed. Many researchers pay attention to this approach. In [18] authors has applied constant plus proportional rate RLC used in [17] for Permanent Magnet Synchronous Motors PMSM drive system. The results showed that, the proposed method gave fast speed response time, with good rejection of disturbances, but the robustness of the system against parametric uncertainties was not considered. In [19] authors propose an improvement of power rate RLC of the article [17], according the distance between the initial states and the sliding surface, the system states can reach the switching surface fast when far away from it, and the speed slow down when the states enter into certain scope. The proposed method gives the system more rapidness and the dynamic system without high frequency chattering has been achieved.

The main aim of this study is to investigate the RLC method in vector-controlled IM for speed and position control system based on new exponential reaching law (ERL) that satisfy the condition of stability according to Lyapunov theorem. This method was applied successfully for robot arm in [20]. This paper is organized as follows; the dynamic model of induction motor and description of field-oriented control are given in Section 2, Details of sliding-mode controllers design is given in Section 4, In Section 5, we have focused interest on the theoretical of the proposed method. This work will be completed by a simulation in Section 6, and finally, some concluding remarks are given in Section 7.

# 2. FIELD ORIENTED CONTROL OF IM

## 2.1 Dynamic Model of Induction Motor

The dynamic model of three-phase, Y-connected induction motor can be expressed in the d-q synchronously rotating frame as [1]:

$$\begin{cases} \frac{di_{sd}}{dt} = -\frac{1}{\sigma L_s}(R_s + \frac{R_r L_m^2}{L_r^2})\, i_{sd} + \omega_s i_{sq} + \frac{1}{\sigma L_s}(\frac{R_r L_m}{L_r^2})\psi_{rd} + \frac{1}{\sigma L_s}(\frac{L_m}{L_r})\omega\psi_{rq} + \frac{1}{\sigma L_s} v_{sd} \\ \frac{di_{sq}}{dt} = -\omega_s i_{sd} - \frac{1}{\sigma L_s}(R_s + \frac{R_r L_m^2}{L_r^2})\, i_{sq} - \frac{1}{\sigma L_s}(\frac{L_m}{L_r})\omega\psi_{rd} + \frac{1}{\sigma L_s}(\frac{R_r L_m}{L_r^2})\psi_{rq} + \frac{1}{\sigma L_s} v_{sq} \\ \frac{d\psi_{rd}}{dt} = \frac{R_r L_m}{L_r} i_{sd} - \frac{R_r}{L_r}\psi_{rd} + \omega_{sl}\psi_{rq} \\ \frac{d\psi_{rq}}{dt} = \frac{R_r L_m}{L_r} i_{sq} - \omega_{sl}\psi_{rd} - \frac{R_r}{L_r}\psi_{rq} \end{cases} \tag{1}$$

where $i_{sd}$, $i_{sq}$, $\psi_{rd}$, $\psi_{rd}$, and $v_{sd}$, $v_{sa}$ are respectively the d-axis and q-axis stator currents, rotor flux linkages, and stator voltages, $R_s$ and $R_r$ are the stator and rotor resistances, $L_s$ and $L_r$ are the stator and rotor inductances, $L_m$ is the mutual inductances between the stator and the rotor winding, $\sigma = 1 - (L_m^2 / L_s L_r)$ is the total leakage factor, $\omega_s$ is the synchronously rotating angular speed, $\omega$ is the electrical angular speed of the rotor, and $\omega_{sl} = \omega_s - \omega$ is the slip frequency. Moreover, the electromagnetic torque $T_e$ can be expressed in terms of stator currents and rotor flux linkages as:

$$T_e = p\frac{3}{2}\frac{L_m}{L_r}(\psi_{rd} i_{sq} - \psi_{rq} i_{sd}) \tag{2}$$

where $p$ is pole pair number.

The motional equation of IM is described as:

$$J\frac{d\Omega}{dt} = T_e - T_L - f_v\,\Omega \tag{3}$$

Where $J$ is the moment of inertia, $T_L$ is the load torque, $f_v$ is the viscous friction coefficient, and $\Omega = \omega / p$ is the mechanical rotor speed.

## 2.2 Description of Indirect Field-Oriented Control

The principle of indirect field-oriented control system of an induction motor is that the d-q coordinate's reference frame is locked to the rotor flux vector, this results in a decoupling of the variables so that flux and torque can be separately controlled by stator direct-axis current $i_{sd}$, and quadrature-axis current $i_{sq}$, respectively, like in the separately excited dc machine. To perform the alignment on a reference frame revolving with the rotor flux requires information on the modulus and position of the rotor flux.

When the reference frame *d-q* is aligned with the rotor flux as in Figure 1, we have:

$$\begin{cases} \psi_{rd} = \boldsymbol{\psi}_r \\ \psi_{rq} = 0 \end{cases} \tag{4}$$

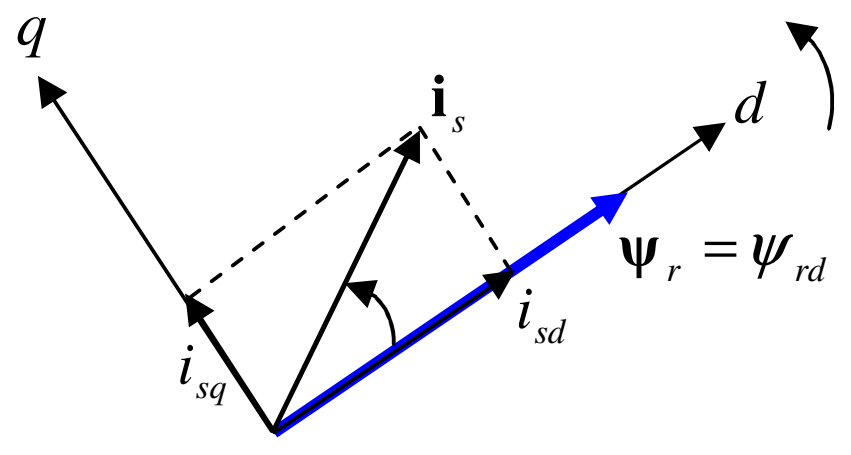

Figure 1. Field orientation in *d-q* reference frame

Where $\boldsymbol{\psi}_r$ , $\mathbf{i}_s$, are the flux and current vectors. Then the system (1) is rewritten as:

$$\begin{cases} \dfrac{di_{sd}}{dt} = -\dfrac{1}{\sigma L_s}\left(R_s + \dfrac{R_r L_m^2}{L_r^2}\right) i_{sd} + \omega_s i_{sq} + \dfrac{1}{\sigma L_s}\left(\dfrac{R_r L_m}{L_r^2}\right) \psi_{rd} + \dfrac{1}{\sigma L_s} v_{sd} \\ \dfrac{di_{sq}}{dt} = -\omega_s i_{sd} - \dfrac{1}{\sigma L_s}\left(R_s + \dfrac{R_r L_m^2}{L_r^2}\right) i_{sq} - \dfrac{1}{\sigma L_s}\left(\dfrac{L_m}{L_r}\right)\omega\, \psi_{rd} + \dfrac{1}{\sigma L_s} v_{sq} \\ \dfrac{d\psi_{rd}}{dt} = \dfrac{R_r L_m}{L_r} i_{sd} - \dfrac{R_r}{L_r}\psi_{rd} \\ \omega_{sl} = \dfrac{L_m}{T_r}\dfrac{i_{sq}}{\psi_{rd}} \end{cases} \tag{5}$$

Where $T_r = L_r/R_r$ is the rotor time constant.

The electromagnetic torque and the synchronous angular speed can be expressed then as:

$$T_e = K_t\, i_{sq} \tag{6}$$

Where : $K_t = p\dfrac{3}{2}\dfrac{L_m}{L_r}\,\psi_{rd}$

$$\omega_s = p\Omega + \frac{L_m}{T_r}\frac{i_{sq}}{\psi_{rd}} \tag{7}$$

The position is determined then as follows:

$$\theta_s = \int \omega_s .dt \tag{8}$$

## 3. SLIDING MODE CONTROLLERS

The transient dynamic response of the system is dependent on the selection of the sliding surfaces. four sliding surfaces are defined, $S_\Omega$ for speed controller, $S_\theta$ for position controller, $S_d$ and $S_q$ for direct and quadratic currents controllers respectively. First, let design the new exponential reaching law.

### 3.1. Design of the Exponential Reaching Law (ERL)

Since the conventional SMC may not handle internal and external uncertainties during the reaching mode, the introduction of exponential approach in IRFOC method guarantees sliding behavior throughout entire response.

In the RLC approach, a difference equation that specifies the dynamic of the switching function is first chosen:

$$\dot{S} = -k\, sign(S) \tag{9}$$

The term $-k$ force the state to steer toward the chosen surface rapidly when $S$ is large in a finite time given by:

$$t_r = \frac{|S_0|}{k} \tag{10}$$

It is obvious that, increasing $k$ leads to a shorter time of reaching mode. However, one must take into consideration when determining the coefficient $k$, because large values induce high frequency oscillations.

According to [20], an exponential term of reaching law is proposed which can adapt to the variations of the switching function. This is given by:

$$\dot{S} = -\frac{k}{N(S)} sign(S), \qquad k > 0 \tag{11}$$

Where: $$N(S) = \delta_0 + (1-\delta_0)e^{-\alpha|S|^P} \tag{12}$$

and: $$\begin{cases} 0 < \delta_0 < 1 \\ p > 0, \quad p \in N \\ \alpha > 0 \end{cases} \tag{13}$$

From (12) it can be seen that, for larger values of $|S|$, $N(S)$ tends $\delta_0$ and then $k/N(S)$ converge to $k/\delta_0$ which is greater than $k$. Otherwise, if $|S|$ decrease, $N(S)$ approaches one, and converge to $k$. This means that, in the first case, the convergence toward the sliding surface will be faster, and in second cases, $k/N(S)$ decreases gradually, which limit the chattering. From this analysis, one can see that, the proposed ERL approach is adaptive to the variations of the switching surfaces.

Moreover, in proposed ERL approach, the time to reach the sliding surfaces is shorter than conventional SMC given in (10), indeed, let $t_r'$ define the reaching time, integrating (11) between zero and $t_r'$ yield:

$$t_r' = \frac{1}{k}\left( \delta_0 |S_0| + (1-\delta_0) \int_0^{S_0} sign(S)\, e^{-\alpha |S|^P} dS \right) \tag{14}$$

For S ≥ 0 and S ≤ 0, (31) can be rewritten as:

$$t_r' = \frac{1}{k}\left( \delta_0 |S_0| + (1-\delta_0) \int_0^{|S_0|} e^{-\alpha |S|^P} dS \right) \tag{15}$$

Subtracting (31) from (31) yields:

$$t_r' - t_r = \frac{(1-\delta_0)}{k}\left( \int_0^{|S_0|} [e^{-\alpha |S|^P} - 1]\; dS \right) \tag{16}$$

Knowing that $(e^{-\alpha |S|^P} - 1) \le 0, \forall S$ , then $t_r' \le t_r$ .

This demonstration shows that, reaching speed in the proposed ERL is increased for the same gain *k*, in other hand, for the same reaching time the proposed approach reduces chattering (because *k* is smaller).

Thus, the control law in Equation (9) may be sensitive to uncertainties and disturbances acting on the system during the reaching phase. Therefore, it may be adjusted to improve the system performance.

For more details on how to choose the ERL parameters see [20].

### 3.2. Design of Speed Controller

We choose integral sliding surface [21]:

$$S_\Omega(t) = e(t) - \int_0^t (k_\Omega - a)\; e(\tau)\; d(\tau) \tag{17}$$

Where $k_\Omega$ is a negative constant.

Considering the mechanical equation we have:

$$\dot{\Omega} + a\Omega + f_1 = b\; i_{sq} \tag{18}$$

Where: $a = \frac{f_v}{J}$, $f_1 = \frac{T_L}{J}$, $b = \frac{K_t}{J}$; and $\dot{(\;)}$ is the derivative.

Now, we consider the previous mechanical equation with uncertainties as follows:

$$\dot{\Omega} = -(a + \Delta a)\ \omega - (f_1 + \Delta f_1) + (b + \Delta b)\ i_{sq} \quad (19)$$

Where the terms $\Delta a$, $\Delta f_1$, $\Delta b$ *are* the uncertainties of the terms $a$, $f_1$ and $b$ respectively.

The error of speed tracking is defined as:

$$e_{\Omega}(t) = \Omega(t) - \Omega^*(t) \quad (20)$$

Where $\Omega^*(t)$, is the reference rotor speed.

The time derivative of (20) gives:

$$\dot{e}_{\Omega}(t) = \dot{\Omega}(t) - \dot{\Omega}^*(t) = -a\ e(t) + u(t) + d(t) \quad (21)$$

Where:

$$u(t) = b\ i_{sq} - a\ \Omega^*(t) - f_1(t) - \dot{\Omega}^*(t) \quad (22)$$

$$d(t) = -\Delta a\ \Omega(t) - \Delta f_1 + \Delta b\ i_{sq}(t) \quad (23)$$

The command *u(t)* is defined by:

$$u(t) = k_{\Omega}\ e_{\Omega}(t) - \beta\ sign\ (S_{\Omega}(t)) \quad (24)$$

Where $\beta$ is the switching gain, it must be chosen so that $\beta \geq |d(t)|\ \forall\ t$, so the speed tracking error $e_{\Omega}(t)$ tends to zero as the time tends to infinity. Substituting (24) in (22) gives the torque current reference:

$$i_{sq}^* = \frac{1}{b}[k_{\Omega}\, e_{\Omega} - \beta\, sign(S_{\Omega}) + a\Omega^* + \dot{\Omega}^* + f_1] \quad (25)$$

### 3.3. Design of Current Controllers

Precise and fast current control is essential to achieve high static and dynamic performance for the IRFOC of induction motors. If the stator currents are not adjusted precisely and with fast dynamics to the command values, cross coupling will appear between the motor torque and rotor flux. Thus, the performance of the IRFOC degrades.

The current controllers design uses two sliding surfaces for d-axis and q-axis stator currents respectively defined as:

$$S_d = i_{sd}^* - i_{sd} \quad (26)$$

$$S_q = i_{sq}^* - i_{sq} \quad (27)$$

where $i_{sd}^*$ , $i_{sq}^*$ are the command values of stator current component. The controller has to maintain the system on the sliding mode always, that is $S_d = S_q = 0$. For this, the direct method of Lyapunov is used in stability analysis.

Considering the Lyapunov function candidate:

$$V = \frac{1}{2} S^2 > 0 \tag{28}$$

Its time derivative is:

$$\dot{V} = S \dot{S} \tag{29}$$

From Lyapunov theorem we know that if $\dot{V}$ is negative definite, the system trajectory will be driven and attracted toward the sliding surface and remain sliding on it until the origin is reached asymptotically.

The control inputs $v_{sd}^*, v_{sq}^*$ are selected as follows:

$$v_{sd}^* = v_{sdeq} + v_{sdn} \tag{30}$$

$$v_{sq}^* = v_{sqeq} + v_{sqn} \tag{31}$$

Is noted that, $v_{sdeq}, v_{sqeq}$ are the equivalent control, it can be estimated from the model parameter and measured states, as follows:

$$v_{sdeq} = \sigma L_s (\dot{i}_{sd}^* - \omega_s i_{sq}) + \left( R_s + \frac{R_r L_m^2}{L_r^2} \right) i_{sd} - \frac{L_m R_r}{L_r^2} \psi_{rd}^* \tag{32}$$

$$v_{sdeq} = \sigma L_s (\dot{i}_{sq}^* + \omega_s i_{sd}) + \left( R_s + \frac{R_r L_m^2}{L_r^2} \right) i_{sq} - \frac{L_m}{L_r} P \Omega \psi_{rd}^* \tag{33}$$

$v_{sdn}$ and $v_{sqn}$ are the discontinuous control law which forces the system to move on the sliding surface.

$$v_{sdn} = k_{id} \, sign \, (S_d) \tag{34}$$

$$v_{sqn} = k_{iq} \, sign \, (S_q) \tag{35}$$

Where $K_{id}$, $K_{iq}$, are a positives constants, representing the maximum controller output required to overcome parameters uncertainties and disturbances [9]. And *sign* is the *signum* function.

If we choose the *sign* function as usually, there will be large chatter in the output. This problem can be remedied by using a continuous function in the sliding surface neighbourhood, as usually doing in the variable structure control, that is:

$$Sat(S) = \begin{cases} \frac{1}{\varepsilon} S & \text{if } |S| \le \varepsilon \\ sign(S) & \text{if } |S| > \varepsilon \end{cases} \tag{36}$$

Where the constant factor ε defines the thickness of the boundary layer and *Sat* is the saturation function.

### 3.4. Position Control of IM

The position error is defined as follows:

$$e_\theta = \theta - \theta^* \tag{37}$$

when the sliding surface is given by:

$$S_\theta = \lambda e_\theta + \dot{e}_\theta \tag{38}$$

The motional equation (18) can be rewritten:

$$\ddot{\theta} = -a\dot{\theta} + b i_{sq} - f_1 \tag{39}$$

Then:

$$\dot{S}_\theta = \lambda\dot{\theta}^* + \ddot{\theta}^* + (a - \lambda)\dot{\theta} - b i_{sq} + f_1 \tag{40}$$

For $\dot{S}_\theta = 0$, we drive the equivalent control as:

$$i_{sqeq} = \frac{1}{b}[\lambda\dot{\theta}^* + \ddot{\theta}^* + (a - \lambda)\dot{\theta} + f_1] \tag{41}$$

The discontinues control is:

$$i_{sqn} = k_\theta \, sign(S_\theta) \tag{42}$$

Where $k_\theta$ is a positive gain. The *sign* function is replaced by *Sat* as in (36).

## 4. Simulation Results

In order to verify the performances of the proposed ERL method the block diagram of the IM drive shown in Figure 2, was simulated in Matlab/Simulink. According to [22], the induction motor used is a three phase, Y connected, four poles, 3 kW, 1415 rpm/s, 220/380V, 6.9A, 50Hz. Electrical and mechanical parameters are, $R_s$=1.84Ω, $R_r$=1.84Ω, $L_s$ =$L_r$ = 0.17 H, $L_m$ = 0.16 H, $J$ = 0.0154 kg.m$^2$, $Cos\varphi$ = 0.89.

The rotor flux is set at it rated values i.e. 0.99Wb, and the electromagnetic torque current command ($i_{sq}^*$) is limited at 7 A, to avoid saturation and overload transients, and the VSI frequency at 5 kHz.

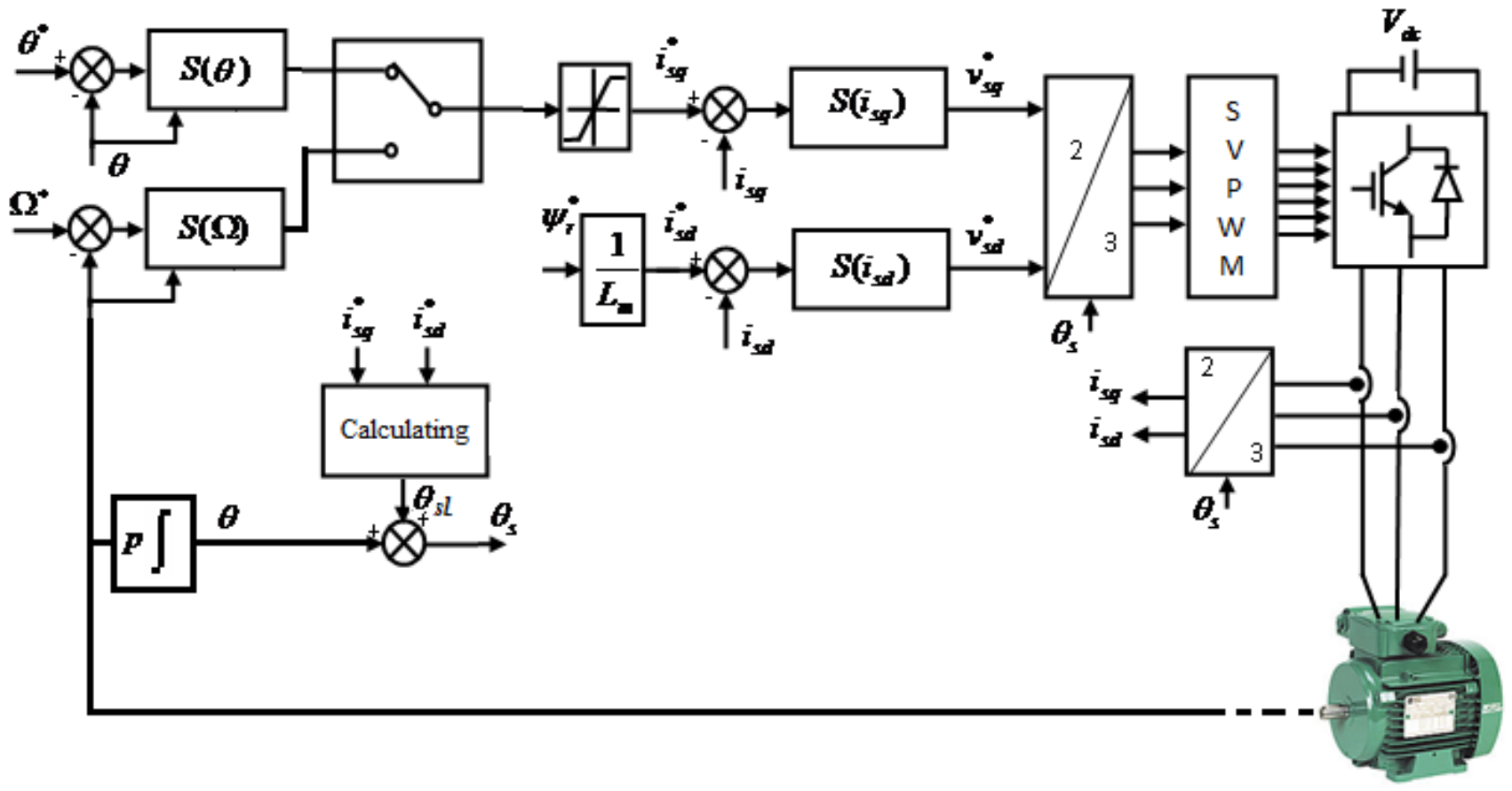

Figure 2. Global scheme of IM control

## 4.1. Speed Performance Analysis

The values of the SMC and ERL parameters are in the following; $k_\Omega$ = -5000, $\beta$=5, $k_{id}$=30, $k_{iq}$=250. $\delta_{0\Omega} = \delta_{0d} = \delta_{0q} = 0.01, \alpha_\Omega = 3, \alpha_d = 3, \alpha_q = 3, P_\Omega = P_d = P_q = 2.$

The test is designed for speed regulation with multiple step commands, starting without load and with disturbance application ($T_L$=10 Nm) at t= 0.65s.

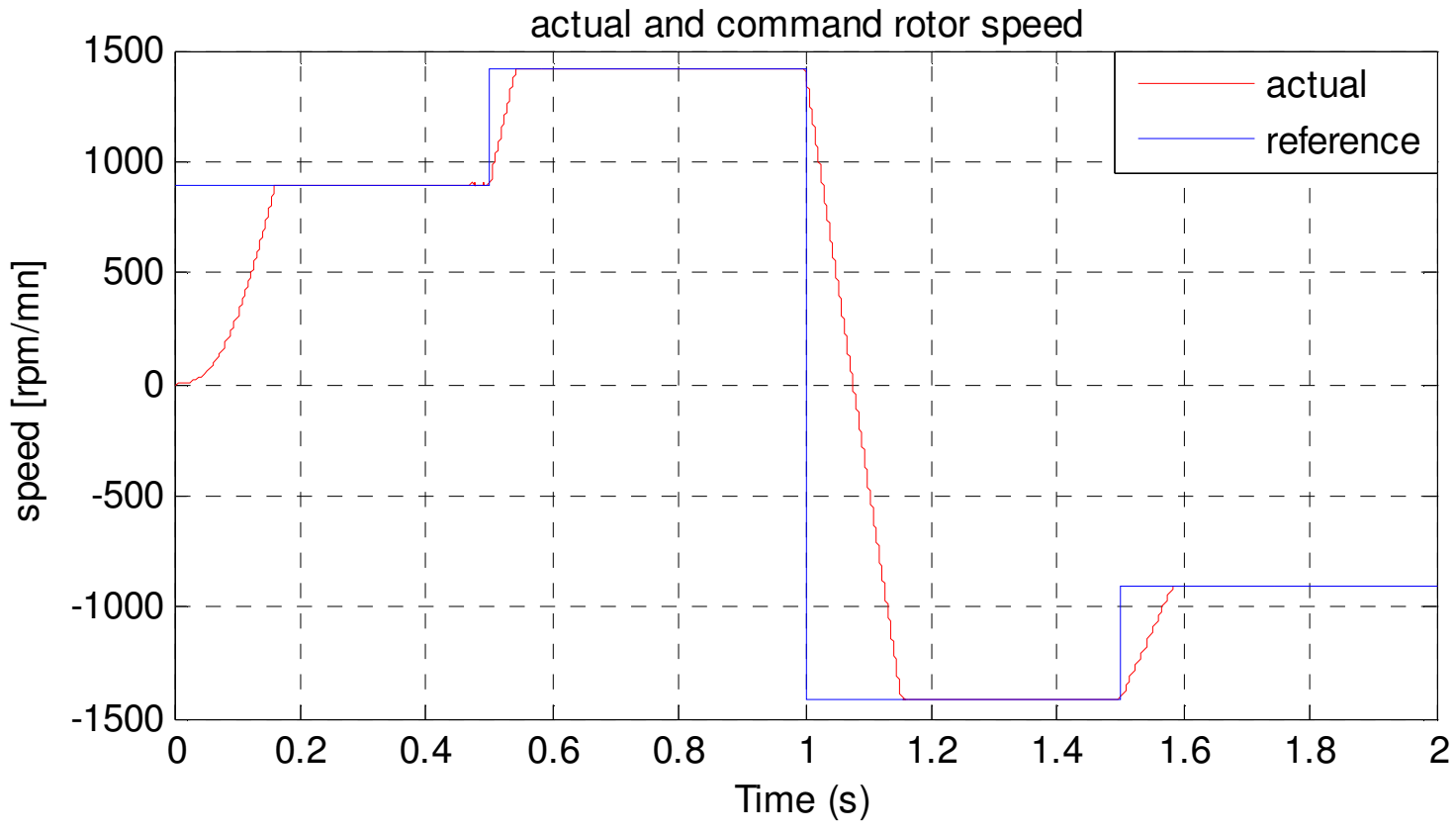

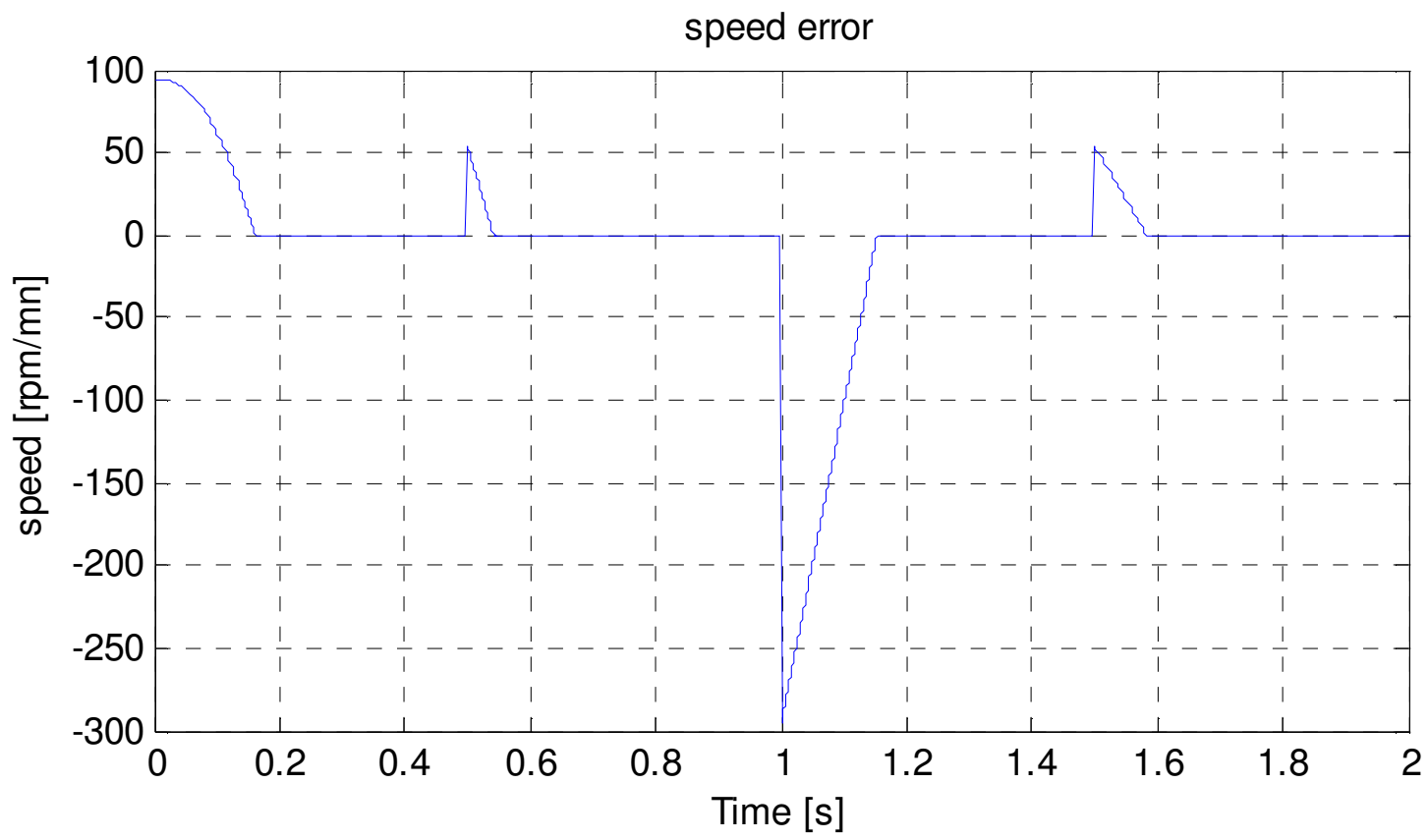

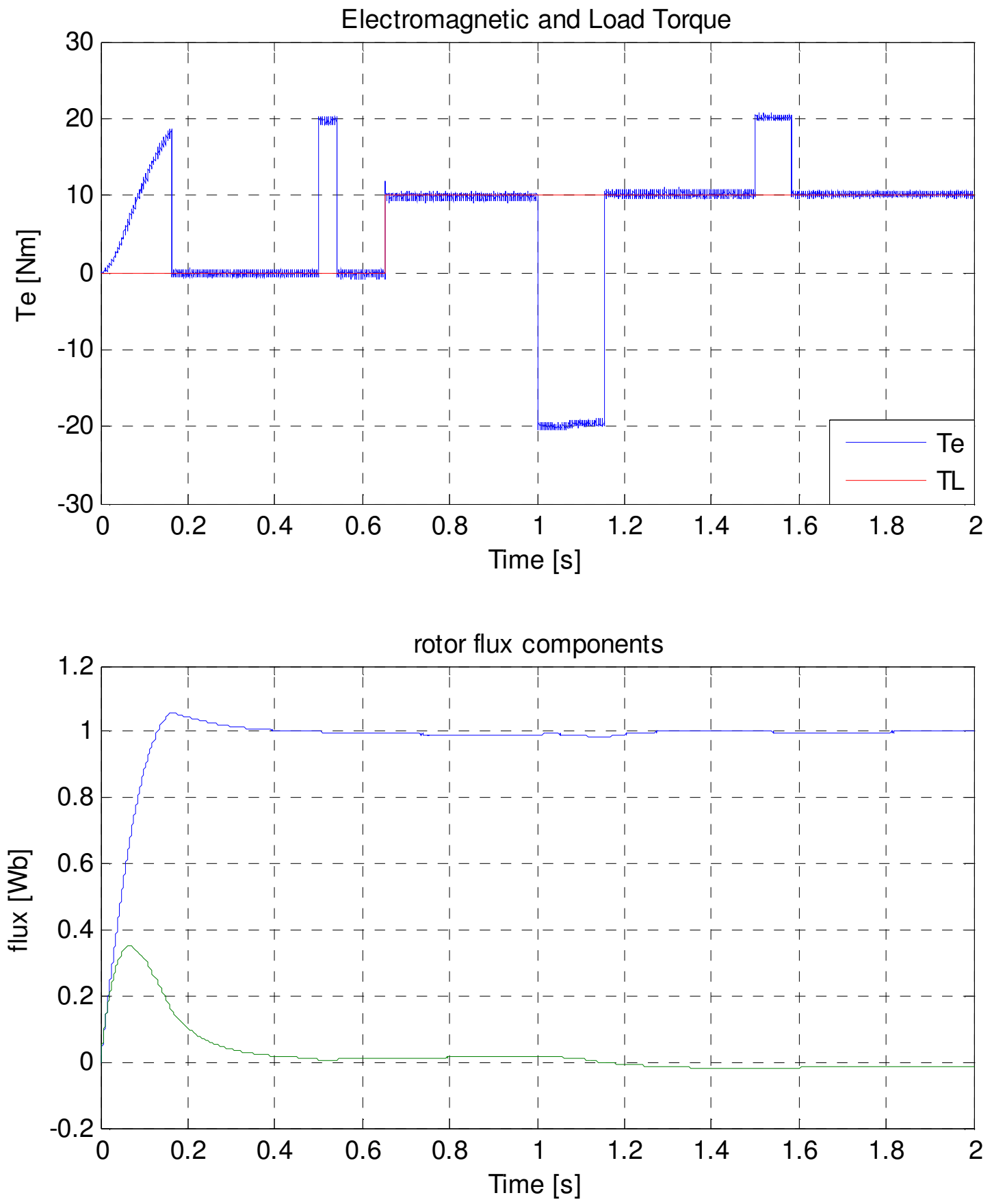

Figure 3. Field orientation in *d-q* reference frame

On the results shown in Figure 3 one can see a high performance speed response obtained despite the disturbance, and the electromagnetic torque rises to its reference instantly after the transients. From the rotor flux waveforms, it is obvious that the ideal IRFOC is achieved in all working conditions.

## 4.2. Position Performance Analysis

The position tracking control is tested with rectangular waveform reference limited at $\theta^* = \pm 240^\circ$, starting at $t=0.5s$ with no-load and when the disturbance is applied at $t=0.6s$. And the SMC parameters are; $k_\theta = 20$, $\lambda=13.85$, $k_{id}=150$, $k_{iq}=150$. The ERL parameters are the same as previous.

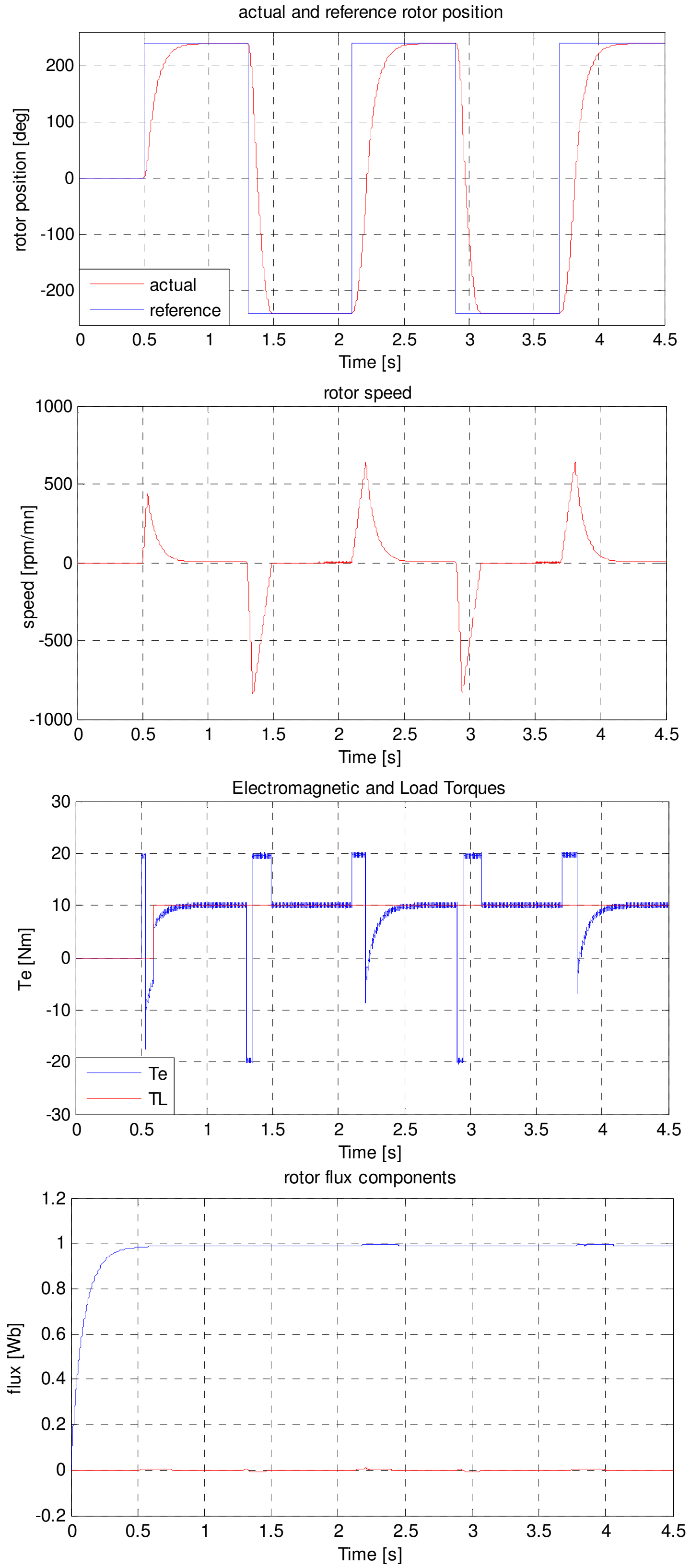

Figure 4. Field orientation in *d-q* reference frame

The results depicted on Figure 4, show a very high performance tracking control without overshoot or steady-state error even after applying of the load torque on the motor shaft. We can also see the rotor flux components which fully maintained the principle of the IRFOC.

## 4.3. Parameters Uncertainties Tests

The robustness of the method is now considered when a step variation about 70% of nominal rotor resistance value is introduced at $t=0.7s$, and for different values on moment of inertia. The Figure 5 and Figure 6, shows that the behavior of the controller is insensitive to these parameters uncertainties for both speed and position control.

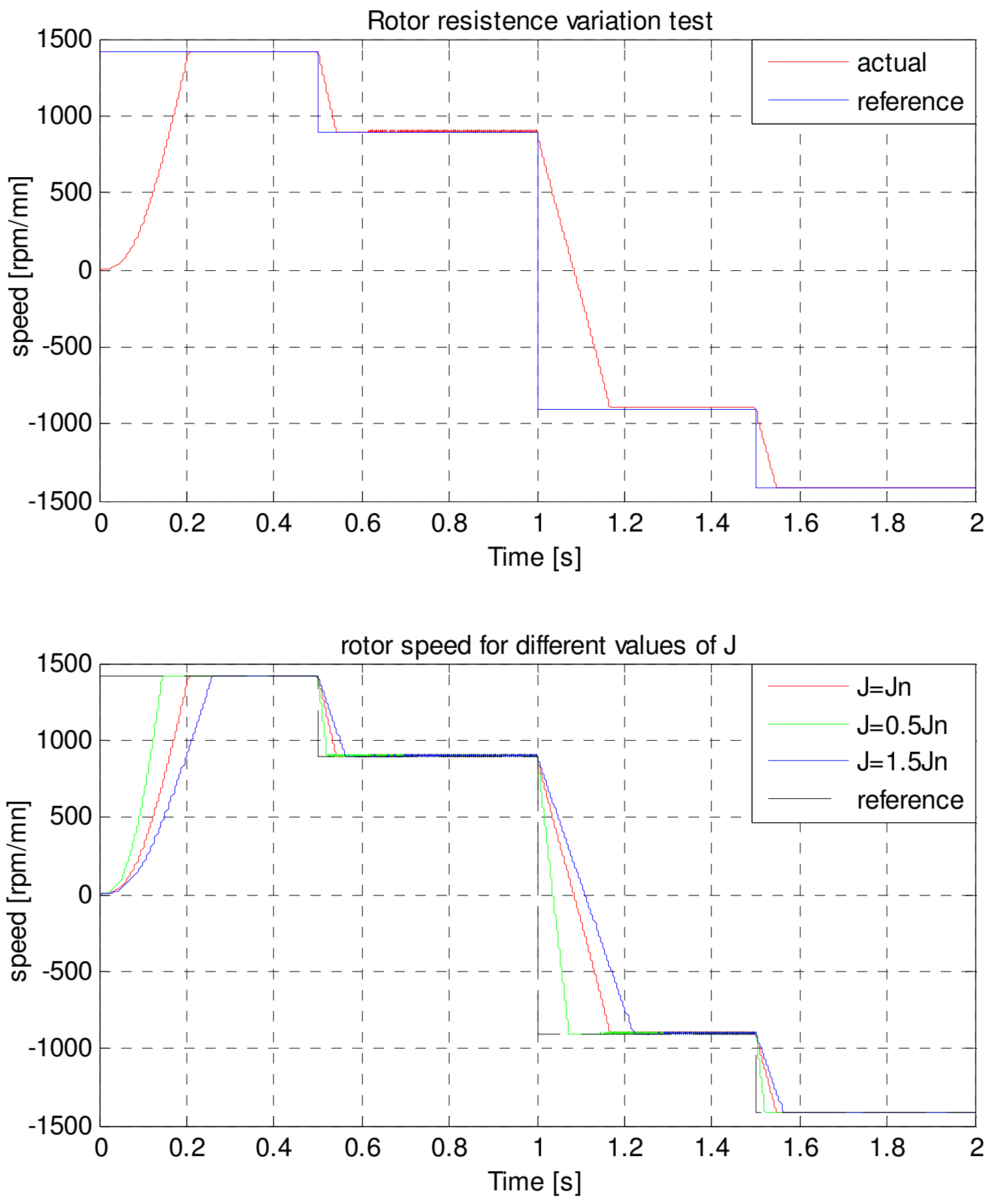

Figure 5. Speed tracking robustness tests

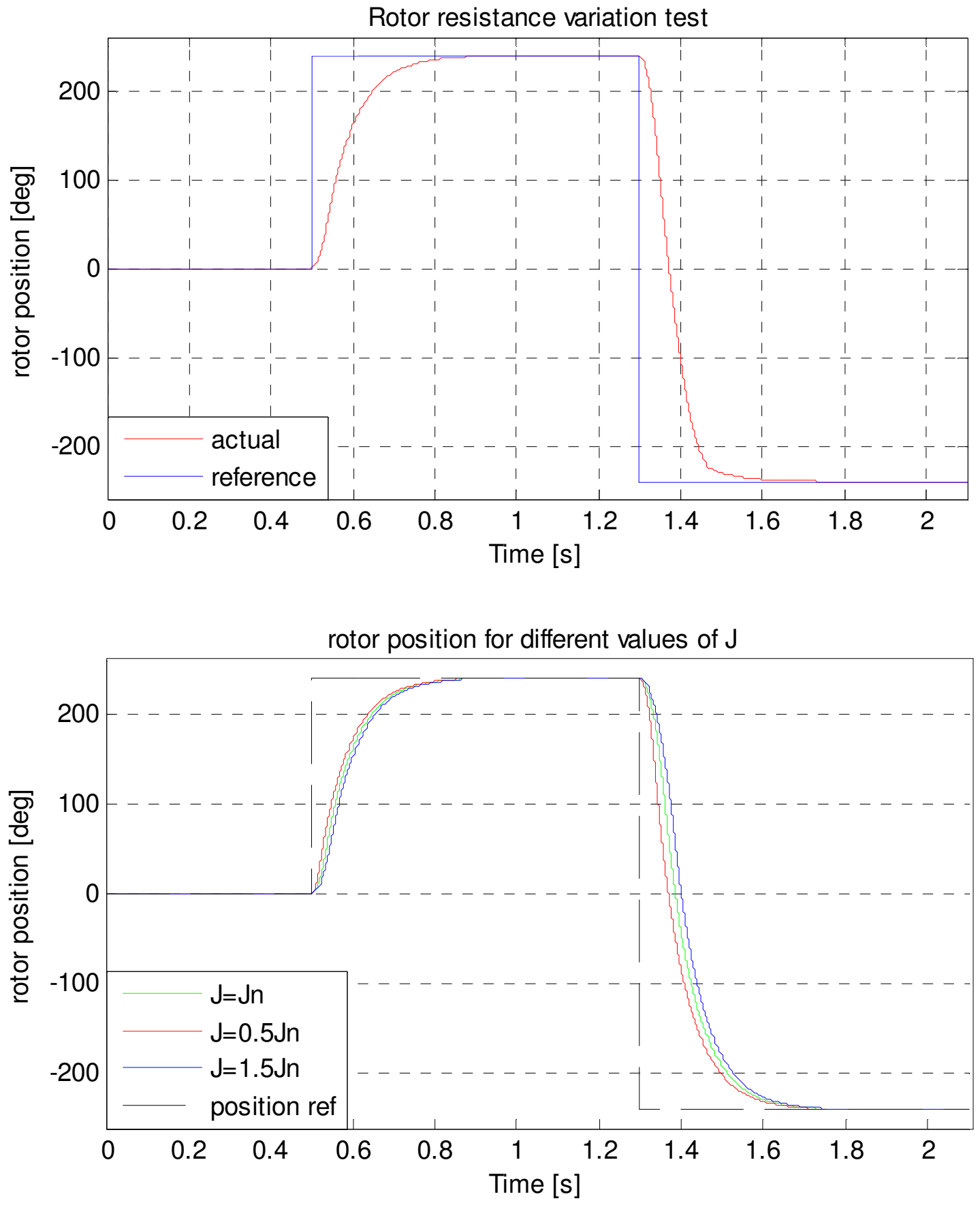

Figure 6. Position robustness tests

## 5. Conclusions

In this study, a new approach of robust IM control was given to improve the discontinuous control law part of a classical sliding mode controller. The results have shown that the proposed method improves the system transient response and lessens the effect of disturbances as it decreases the reaching time. The distinctive feature of the scheme compared to conventional SMC, is its robustness and high tracking performance to different commands values and parameter mismatch for both position and speed control without chattering suffering. The simulation results confirm the effectiveness of the proposed ERL control approach.

## References

[1] B. K. Bose, "Modern Power electronics and AC drives," The University of Tennessee, Knoxville, USA, Prentice Hall, 2002.

[2] H. van der Broeck, H.C. Skudelny, G.V. Stanke, "Analysis and Realization of a Pulse Width Modulation Based on Voltage Space Vectors," IEEE Trans, Ind. App. vol 24, pp 142-150, Jan/Feb 1987.

[3] Fnaiech, M.A., Betin, F., Capolino, G.-A., Fnaiech, F., "Fuzzy Logic and Sliding-Mode Controls Applied to Six-Phase Induction Machine With Open Phases," IEEE Transactions On Industrial Electronics, Vol. 57, No. 1, january 2010.

[4] Uddin, M.N.; Hao Wen, "Development of a Self-Tuned Neuro-Fuzzy Controller for Induction Motor Drives," IEEE Trans. On Industry Applications, Vol. 43, No. 4, pp. 1108-1116, July-Aug. 2007.

[5] Chih-Min Lin, Chun-Fei Hsu, "Neural-network-based adaptive control for induction servomotor drive system," IEEE Trans. On Industrial Electronics, Vol. 49, No. 1, pp. 115-123, Feb.2003.

[6] K. Jamoussi, M. Ouali, H. Charradi, "A Sliding Mode Speed Control of an Induction Motor," Am. J. Applied. Sci. 4(12): pp 987-994, Science Publications 2007.

[7] Vadim I. Utkin, "Adaptive Sliding-Mode Neuro-Fuzzy Control of the Two-Mass Induction Motor Drive Without Mechanical Sensors," IEEE Trans. Ind. Electron. Vol. 57, No. 2, pp. 553–564, Feb. 2010.

[8] Asif Sabanovic, "Variable Structure Systems With Sliding Modes in Motion Control—A Survey," IEEE Trans. Ind. Inf. vol 7, No.2 pp. 212–223, May 2011.

[9] Vadim I. Utkin, "Sliding mode control design principles and applications to electric drives," IEEE Trans. Ind. Electron. vol 40, pp. 23–36, 1993.

[10] Zhimei Chen, Wenjun Meng, Jinggang, He Wang, "Fuzzy Reaching Law Sliding Mode Control of Robot Manipulators," IEEE Pacific-Asia Workshop on Computational Intelligence and Industrial Application, pp. 393-397, 2008.

[11] Hoon Lee, Vadim I. Utkin, "Chattering suppression methods in sliding mode control systems," Annual Reviews in Control 31, pp. 179-188, Elsevier Ltd. 2007.

[12] J. C. Hung, "Chattering Handling for Variable Structure Control Systems," Proceedings of the IECON, International Conf. on vol. 3, pp. 1968-1972, Nov 1993.

[13] L. K. Wong, F. H. F. Leung, P. K. S. Tam, "A chattering elimination algorithm for sliding mode control of uncertain non-linear systems," Mechatronics 8, pp. 765-775, Elsevier Ltd. 1998.

[14] E. Iglesias, Y. García, M. Sanjuan, O. Camacho, C. Smith, "Fuzzy surface-based sliding mode control," ISA Transactions, Volume 46, Issue 1, pp. 73-83, Feb. 2007.

[15] Damiano, A., Gatto, G.L., Marongiu, I., Pisano, A., "Second-Order Sliding-Mode Control of DC Drives," IEEE Trans. Ind. Electron. Vol. 51, No.2, pp. 364–373, 2004.

[16] Traore, D., Plestan, F., Glumineau, A., de Leon, J., "Sensorless Induction Motor: High-Order Sliding-Mode Controller and Adaptive Interconnected Observer," IEEE Trans. On Ind. Electron. Vol. 55, No.11, pp. 3818 - 3827, Nov. 2008.

[17] Weibing Gao; Hung, J.C., "Variable Structure Control of Nonlinear Systems: A New Approach," IEEE Trans. On Ind. Electron. Vol. 40, No.1, pp. 45 - 55, Feb. 1993.

[18] Ying Liu, Bo Zhou, Haibo Wang, Sichen Fang. "A New Sliding Mode Control for Permanent Magnet Synchronous Motor Drive System Based on Reaching Law Control," 4th IEEE Conf. Industrial Electron. And Appl., pp. 1046– 1050, ICIEA 2009.

[19] Shibin Su, Heng Wang, Hua Zhang, Yanyang Liang, Wei Xiong, "Reducing Chattering Using Adaptive Exponential Reaching Law," Sixth International Conference on Natural Computation (ICNC), Vol.6, pp. 3213-3216, 2010.

[20] Fallaha, C.J., Saad, M., Kanaan, H.Y., Al-Haddad, K., "Sliding-Mode Robot Control With Exponential Reaching Law," IEEE Trans. On Ind. Electron. Vol. 58, No.2, pp. 600 - 610, Feb. 2011.

[21] Barambones O., Garrido A.J., Maseda F.J. "A Robust Field Motor with Flux Oriented Control of Induction Observer and Speed Adaptation," Emerging Technologies and Factory Automation, IEEE Conference, vol.1. pp.245– 252, September. 2003.

[22] Marek Jasiński, "Direct Power and Torque Control of AC/DC/AC Converter-Fed Induction Motor Drives," Ph.D. Thesis (Faculty of Electrical Engineering) Warsaw, Poland, 2005.

## Authors

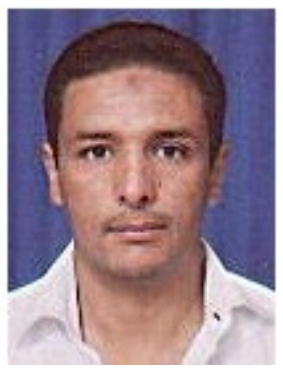

**Salah Eddine Rezgui** was born in Constantine, Algeria in 1971. He received the B.T., L.T., and magisterial degrees in electric machines drives, from the Mentouri Constantine University in 2009. Now he is PhD student and researcher's member in Constantine Electrotechnics Laboratory, His research interests include modern theory application control, electric machines drives, and power electronics.

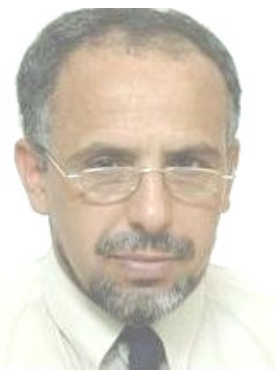

**Hocine Benalla** was born in Constantine, Algeria in 1957. He received the B.S., M.S., and Doctorate Engineer degrees in power electronics, from the National Polytechnic Institute of Toulouse, France, respectively in 1981, 1984. In 1995, he received the Ph.D. degree in Electrical Engineering from University of Jussieu-Paris VI, France. Since 1996, he is with the department of Electrotechnics, at Constantine University Algeria, as a **Professor**. His current research field includes Active Power Filters, PWM Inverters, Electric Machines, and AC Drives.